%% file: main.tex
\documentclass{article}

\usepackage[final,nonatbib]{neurips_2024}

\input{preamble}

\usepackage[utf8]{inputenc} 
\usepackage[T1]{fontenc}    
\usepackage{url}            
\usepackage{amsfonts}       
\usepackage{nicefrac}       
\usepackage{microtype}      

\usepackage{graphicx}
\usepackage{booktabs}

\usepackage[accsupp]{axessibility}  

\usepackage{orcidlink}
\newcommand{\eg}{\emph{e.g., }}

\usepackage{hyperref}

\title{SA-VIS: Sparse frame Annotations for training Video Instance Segmentation}

\author{Edoardo Mello Rella\Mark{1} Ajad Chhatkuli\Mark{1,3} Shipra Jain\Mark{2} Ender Konukoglu\Mark{1} Luc Van Gool\Mark{1,3} \\
\Mark{1}\,CVL, ETH Zurich \Mark{2} Align Technology \Mark{3}\,INSAIT, Sofia University "St. Kliment Ohridski"\\}

\begin{document}
\maketitle
\input{sec/0_abstract}    
\input{sec/1_intro}
\input{sec/2_relatedwork}

\input{sec/3_method}

\input{sec/4_experiment}

\input{sec/5_limitations}
\input{sec/6_conclusion}

\bibliographystyle{splncs04}
\bibliography{main}

\end{document}

%% file: preamble.tex
\usepackage{colortbl}
\usepackage{multirow}
\usepackage{multicol}
\usepackage{amssymb}
\usepackage{pifont}

\newcommand{\cmark}{\ding{51}}

\newcommand{\Mark}[1]{\textnormal{\textsuperscript{#1}}}
\newcommand{\rulesep}{\unskip\ \vrule\ }

\definecolor{mygray}{gray}{0.85}

%% file: sec/0_abstract.tex
\begin{abstract}

Recent online video instance segmentation (VIS) methods have achieved impressive results, thus becoming the preferred approach to segment instances in videos. Despite the resurgence of impressive single image models, the online (or semi-online) VIS approaches outperform single-image models (\emph{e.g.}, based on SAM) by using long sequences of densely annotated frames during training. However, such a training setup of VIS is expensive in the sense of compute as well as dense annotations required. In order to solve these major flaws, we argue that the effective modeling of the instances and their evolution in videos do not require densely annotated frames.
To that end, we propose a simple and effective module, called Past-frames Feature Propagation (PFP) which aggregates low-dimensional features from the image encoder of multiple frames. This simple low-compute module provides tremendous learning capability in using sparse video frame labels for end-to-end training. Combined with a light-weight frame-specific Instance Queries, our Sparse frame Annotation VIS (SA-VIS) significantly improves performance over its baseline.
Most interestingly, our simple design that avoids complexities effectively bridges the gap in accuracy between training on sparsely and densely annotated video sequences. This translates to a mere $0.4\%$ drop in performance of SA-VIS when using annotations for only $1/5$ of the images in the dataset
Empirically, SA-VIS shows strong improvements over the baseline on YouTube-VIS 2019/2021/2022 and Occluded VIS (OVIS) and an over $1\%$ improvement in $AP$ on the state-of-the-art in a limited annotations scenario.

\end{abstract}

%% file: sec/1_intro.tex
\section{Introduction}
\label{sec:intro}
\input{figs/fig_tex/teaser}

Video Instance Segmentation (VIS) \cite{Yang_2019_ICCV} consists of segmenting and tracking all the countable instances of objects in a video sequence. It has many potential applications ranging from autonomous driving to video analytics, video editing and augmented reality. However, it is an inherently complex and difficult task, requiring identification of an uncertain number of potentially occluded and/or overlapping instances of different sizes in videos of varying length/speed. The complexity of the task is also pronounced in the densely annotated video sequences required to train a robust State-of-the-Art VIS model. However, large-scale video datasets are extremely expensive to annotate, given the number of frames in each video. Solutions for VIS that are robust to fewer annotation labels and lower memory/compute training, although potentially having many implications, have garnered little attention.

Most of the present-day methods can be categorized as online or offline. Online models \cite{IDOL, hannan2023gratt, ying2023ctvis, Yang_2021_ICCV, huang2022minvis, Choudhuri_2023_CVPR} formulate predictions by processing the frames one at a time, as if the frames were provided \emph{online} by a camera. Conversely, offline models \cite{wu2021seqformer, wang2020end, Jiang2023STC, maskfreevis, VITA, Cheng2021Mask2FormerFV, hwang2021ifc, Zhang_2023_ICCV} process the entire (stored) video simultaneously, dividing it into chunks if necessary. Despite the different use of the video data, recent online models \cite{IDOL, hannan2023gratt, ying2023ctvis} have been able to achieve impressive results and even outperform the offline counterparts. Online models achieve superior results by accurately detecting and representing instances on a frame level, and then associating them across frames either based on feature similarity \cite{IDOL, ying2023ctvis} or query propagation \cite{hannan2023gratt, Choudhuri_2023_CVPR}.

Traditional online models \cite{IDOL, huang2022minvis, han2022visolo, Yang_2021_ICCV} typically have smaller annotation requirements as they only process single or couple of frames rather than entire clips. Some recent methods, however, have adapted the offline training approach - still being fully online during inference - of jointly optimizing prediction of multiple frames, \eg 8 frames simultaneously \cite{ying2023ctvis, GenVIS} onto online models with impressive results. The improvement here can also be alluded to the result of better inductive prior of video learned through training of multiple frames at once. However, the approach requires long sequences of densely annotated data and higher memory/compute for training, thus making it not suitable in many real-world scenarios.

We argue that the use of multiple past frames in a video can provide useful context, even when the past frames are not annotated, with a minimal design.
In an attempt at this approach, we propose to exploit the feature extraction layers of an online model to generate compressed image representations, which can then be stored and re-used to augment the available information when processing a subsequent annotated frame. Such an approach taps in the power of encoder to learn attention between frames, while at the same time uses minimal compute/memory for generating the added compressed frame features. Additionally, we use the rich feature representation created for a frame to generate distinctive queries that further help improve performance.
We note that at the training stage the model computes the loss only on pairs of sparse annotated frames, while using the "frozen" model to obtain the features for the past non-annotated frames.

Overall, our contributions can be summarized as follows:

\begin{itemize}
\item We build an online VIS model that can be trained on pairwise frames from sparsely annotated video sequences by using the past frames features, thus improving the matching of instances across frames.
\item We design a network architecture able to effectively include multiple frames in a video through past-frames feature propagation (PFP) and to formulate individual predictions for each frame through frame-specific instance (FSI) queries.
\item With extensive evaluation, we show that our Sparse frame Annotation for VIS (SA-VIS) achieves state-of-the-art results, and evaluate the contribution of each individual element on performance.
\end{itemize}

%% file: figs/fig_tex/teaser.tex
\begin{figure}[!h]
\centering
\includegraphics[width=0.43\linewidth]{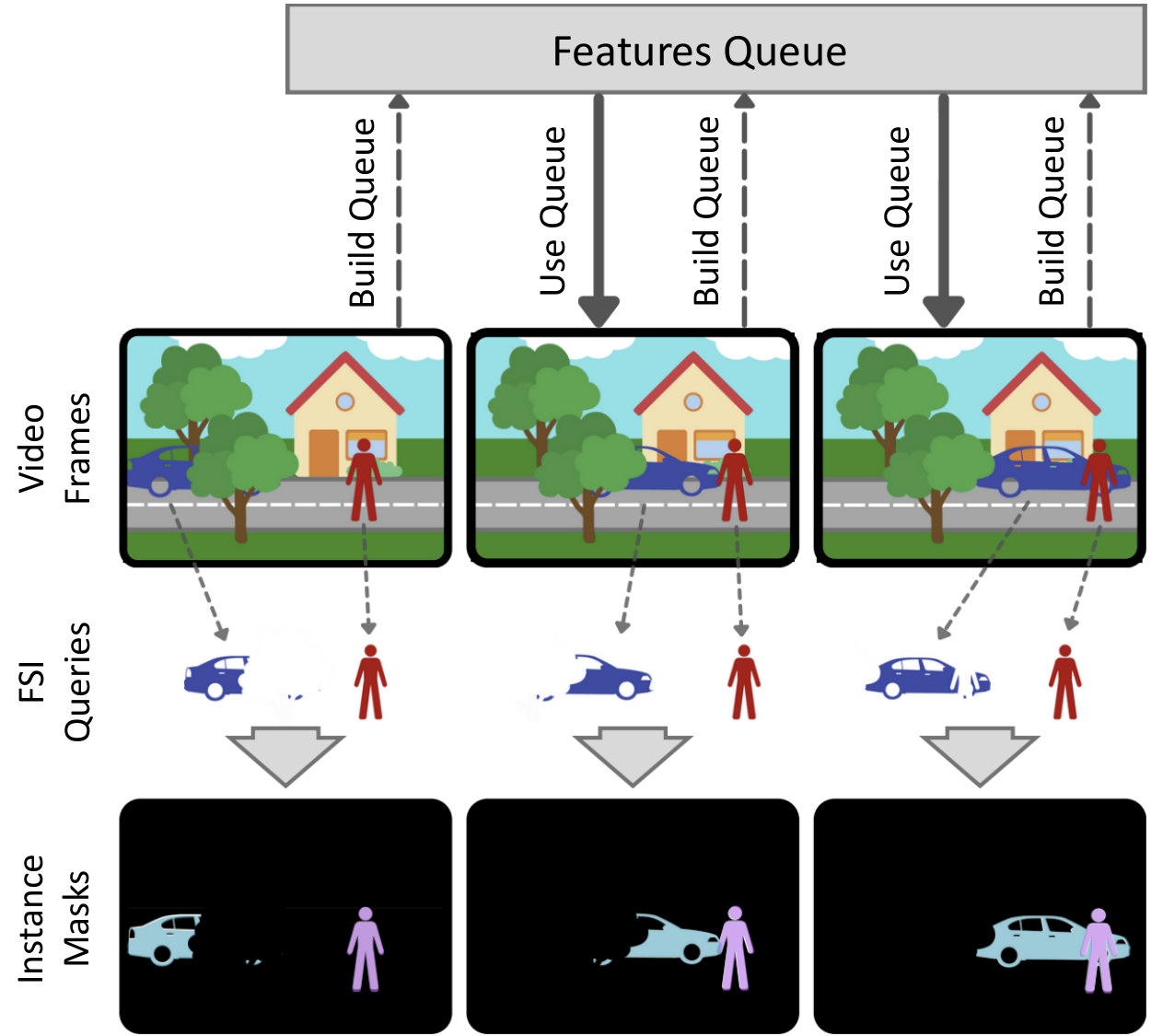}
\hspace{0.015\linewidth}
\rulesep
\hspace{0.015\linewidth}
\includegraphics[width=0.43\linewidth]{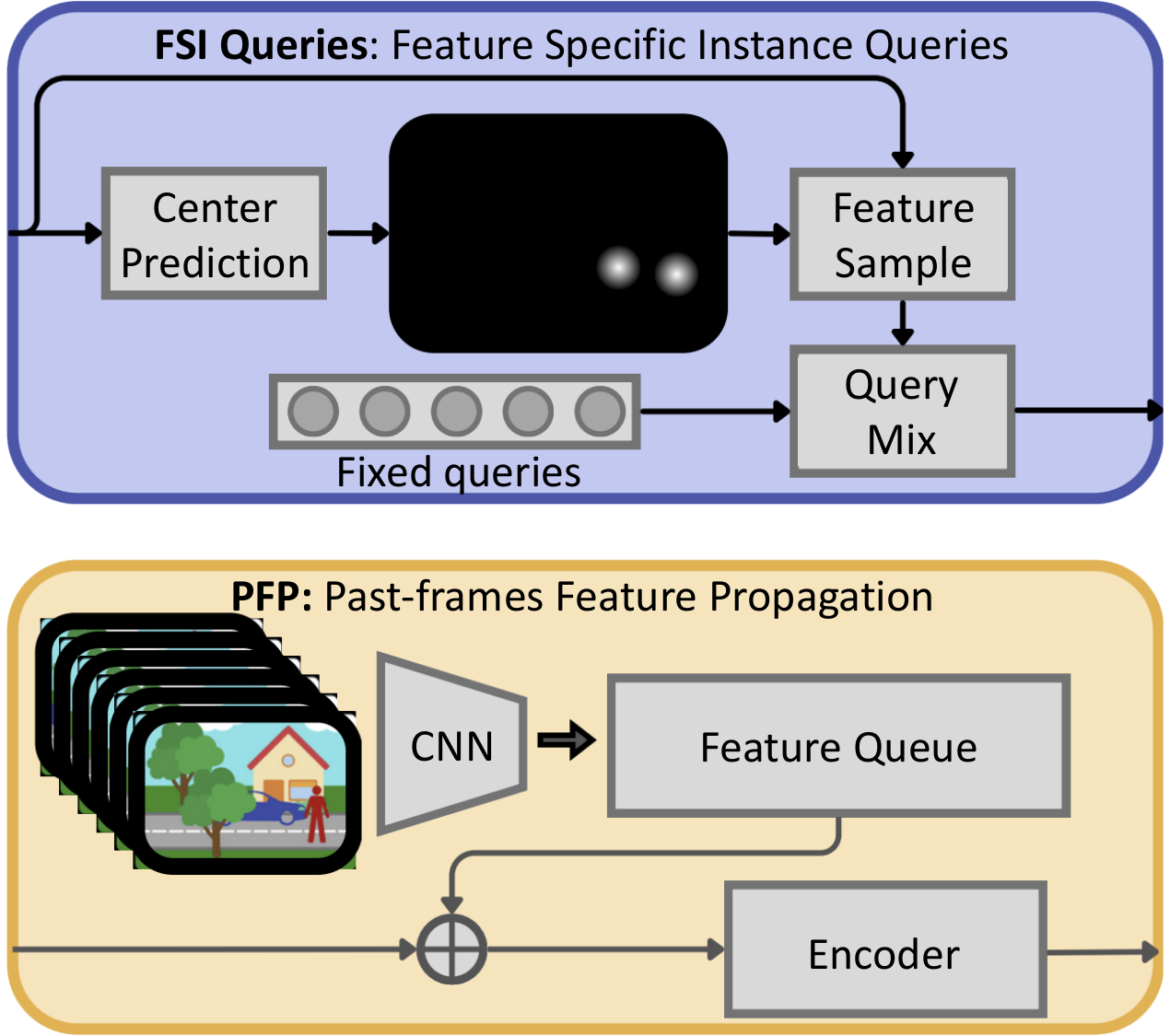}
\caption{\textbf{SA-VIS:} we propose a method that includes \textbf{past frames awareness} and generates \textbf{frame-specific instance queries} for the task of video instance segmentation. Thanks to the addition of the shown modules, we propose a method that \textit{i}) generates queries based on the objects actually visible in the image (\textbf{FSI Queries}), and \textit{ii}) uses the past frames to build useful and lightweight contextual information (\textbf{PFP}). The proposed modules improve the overall accuracy and allow SA-VIS to train with sparsely annotated datasets.}
\label{fig:teaser}
\end{figure}

%% file: sec/2_relatedwork.tex
\section{Related Work}
\label{sec:relwork}

Following common practice we categorize VIS method as either \emph{online} or \emph{offline} and give an overview of relevant works in the respective classes.

\paragraph{Online VIS Methods:} 
As the first online model, which was proposed together with the VIS task, MaskTrack R-CNN \cite{Yang_2019_ICCV} built on top of Mask RCNN \cite{He_2017_ICCV}, by adding a tracking branch able to associate instances across video frames. A number of subsequent works followed on the approach \cite{Voigtlaender_2019_CVPR, porzi2020learning, lin2020video, cao2020sipmask, fu2021compfeat, wu2021track, liu2021sg} mostly by improving tracking. More recently, the use of query-based detectors and segmentors \cite{zhu2021deformable, cheng2021mask2former} and, consequently, the possibility to match instances based on the query embeddings has allowed significant improvements. Most notably, IDOL \cite{IDOL} uses contrastive learning and optimal transport to learn discriminative instance embeddings. Concurrently, MinVIS \cite{huang2022minvis} achieved competitive results without requiring video-based training thanks to bipartite matching for instance association. CTVIS \cite{ying2023ctvis} further improves the results by mimicking at training time the inference scheme through the creation of a memory-bank for instance association; this, though, requires joint training on long annotated video sequences. Following a different approach, several works \cite{zhan2022robust, koner2022instanceformer, hannan2023gratt, Choudhuri_2023_CVPR} achieve instance association by propagating queries across frames. Among the others, GrattVIS \cite{hannan2023gratt} proposes to use a gated residual connection followed by masked self attention which improves propagation on a longer temporal range.

\paragraph{Offline VIS Methods:}
These models \cite{wu2021seqformer, wang2020end, Jiang2023STC, maskfreevis, VITA, Cheng2021Mask2FormerFV, hwang2021ifc, Zhang_2023_ICCV} take entire videos or clips as input and formulate instance segmentation as a single task. Afterwards, they merge the clips from the video - which often have a number of overlapping frames - thus forming entire instance tracklets. VisTR \cite{wang2020end} uses \cite{carion2020end} predicts clip-level masks trajectories in an end-to-end manner from a set of clip-level instance features. IFC \cite{hwang2021ifc} then reduces the computational complexity by designing a lighter inter-frame communication method. SeqFormer \cite{wu2021seqformer} further improves the results by aggregating spatio-temporal context and predicting masks using queries; similarly, also Mask2Former-VIS \cite{Cheng2021Mask2FormerFV} uses a transformer architecture \cite{NIPS2017_3f5ee243} and query-based inference. VITA \cite{VITA} associates instances purely based on object tokens across frames without needing image-level spatio-temporal features. With longer videos in the datasets, a new challenge of offline models is associating instance tracklets from separate clips. Most approaches involve heuristics or overlapping computation, while EfficientVIS \cite{wu2022efficient} allows tracklet linking in an end-to-end learnable way. GenVIS \cite{GenVIS} further changes the approach by strongly reducing the clip length and associating instances through query propagation.

%% file: sec/3_method.tex
\section{Method}
\label{sec:method}

We first explain the neural network architecture for VIS and then the training and inference protocols.
After briefly reviewing the baseline network structure in Section \ref{sec:inst_arch}, Sections \ref{sec:past} and \ref{sec:inst_q} detail the architectural contributions of our work, which allow to successfully include sequences of non-annotated frames. Finally, in Section \ref{sec:train_inf}, we describe the training and inference process with SA-VIS.

\subsection{DeformableDETR for Instance Segmentation}\label{sec:inst_arch}

Consistent with online VIS methods, our base network processes each frame independently. We use a starting network structure based on IDOL \cite{IDOL}. From an input image $I \in \mathbb{R}^{3 \times H \times W}$, a CNN backbone \cite{He_2016_CVPR} extracts multi-resolution features. DeformableDETR \cite{zhu2021deformable} then takes the multi-resolution features together with their fixed positional encoding \cite{carion2020end} and applies multi-resolution self attention to generate a set of features $M$. Then, $N$ learnable object queries are used to decode a set of embeddings $E \in \mathbb{R}^{N \times C}$ that correspond to $N$ predicted instances. From the embeddings, $N$ class labels and bounding boxes are predicted directly using separate 3-layer feed-forward network (FFN). Following \cite{IDOL}, we set $N = 300$. For mask predictions, instead, we use a dynamic mask head \cite{tian2020conditional} that generates instance masks at $1/8$\emph{th} of the input image resolution. On top of this instance segmentation architecture, we add a contrastive embed head using an FFN applied to the instance embeddings $E$ similarly to \cite{IDOL}.

\subsection{Past-frames Feature Propagation (PFP)}\label{sec:past}

Consecutive frames in a video, often display scenes that are highly redundant. This is a possible reason why offline-inspired models~\cite{ying2023ctvis,GenVIS} achieve better performance by using lightweight inter-frame communication, rather than processing frames together at an image level \cite{wang2020end, hwang2021ifc, wu2021seqformer}. On the contrary, online models have to rely on learning discriminative instance features from single frames; despite the recent progress, the task remains challenging. Logic, human experience, and previous works \cite{jin2017video, gadde2017semantic, nilsson2018semantic} suggest that observing the temporal evolution should strongly simplify the task of tracking objects in a video; we argue this is the case also when it is not possible to train the predictions on these frames. Motivated by these reasons, in SA-VIS, we propose to include PFP to improve instance matching while avoiding excessive computation. While in other tasks, feature propagation methods have been proposed \cite{yang2022decoupling, zhang2020unsupervised}, they are often task-specific and complex. Others~\cite{he2022inspro, wu2023onlinerefer} have proposed a complex query propagation with RoIAlign scheme, without relationship to sparse annotations. In contrast, we propose a simple and lightweight feature propagation approach that, without bells and whistles, significantly improves performance. Moreover, it only adds minimal training complexity and data requirements, in contrast to many recent methods trained on densely annotated data.

From each frame $j$, the backbone extracts feature maps at $4$ different resolutions, with the features at level $i$ being $F_i \in \mathbb{R}^{C_i \times H_i \times W_i}$. To propagate information from past frames to the current frames, we use the deepest feature $F_4 \in \mathbb{R}^{2048 \times H / 32 \times W / 32}$. We then use adaptive average pooling on $F_4$ to obtain a $6 \times 6$ feature map; this is followed by an FFN to obtain vectors of size $256$ from $2048$. Finally, positional encoding is added to each vector and the spatial dimensions of the feature map are flattened. This set of image feature vectors $H_j \in \mathbb{R}^{36 \times 256}$ is collected in a Feature Queue (FQ) and constitutes the information from each frame $j$ that is used to provide PFP to the remaining part of the sequence. When processing the $j$th frame, a set of $T$ FQ vectors obtained from the $j-T$th to the $j-1$th frame (i.e. $H_{j-T}, \hdots, H_{j-1}$) and the current $H_j$ are concatenated on their temporal axis and their temporal positional encodings are summed. We then feed them to a 2-layer transformer encoder~\cite{NIPS2017_3f5ee243} comprised of standard multi-head attention layers - computing attention between the vectors $H_i$ - and residual connections. $T$ can be adjusted depending on the application and computation constraints. However, we keep it constant, with $T=20$ in favor of simplicity. The embedding obtained is then combined with the multi-resolution features from the current frame inside the transformer encoder. This is done by inserting a cross-attention layer between each of the self-attention layers of the transformer encoder. This results in richer features $M$ as output of the transformer encoder (refer to Section \ref{sec:inst_arch} for the notation).

We note that the proposed structure is general and may be applied on top of other existing online VIS methods.
The additional module adds little to the model complexity, thanks to the small dimension of the FQ vectors and the minimal number of additional layers.
Furthermore the added compute can be limited by reducing the number of past frames $T$ kept in the FQ.
The overall computational complexity of the method can be seen in Tab. \ref{tab:perf_reb}, which highlights the small overhead of the additions.

\input{tables/ytvis192122}
\input{tables/ovis}

\subsection{Frame-specific Instance (FSI) Queries}\label{sec:inst_q}

Object queries are an important part of most transformer-based instance segmentation and detection methods \cite{carion2020end, zhu2021deformable, cheng2021mask2former, cheng2021maskformer}.
It is believed that these encode positions in the image grid of the object they tend to represent \cite{carion2020end,cheng2021maskformer,choudhuri2023context}. They are more effective than previously used feature proposal networks as only a few hundred queries can represent each object in every image.
However, the performance changes observed when varying the number of queries \cite{cheng2021maskformer} suggest that their use can be improved further.
For VIS, we propose to supplement them with a set of new FSI queries, which are few and unique to the instances that appear in the current frame. This adds a set of queries which are customized to each frame and can thus help the model to achieve higher accuracy.

In order to generate FSI queries, we first detect the instances in the current frame using a lightweight detector, extract a feature vector for each of them and use the feature vector to generate new queries. Following prior work \cite{Cheng_2020_CVPR}, we detect object instances using a single channel heatmap trained to predict Gaussians located at the center of the instances. To integrate this lightweight detector in the network structure described in Sec. \ref{sec:inst_arch}, the multi-resolution features $F_i$ are processed by a set of convolutional layers, bilinearly upsampled and concatenated to obtain a feature map $F_C$. $F_C$ can then be used to predict a single-channel center heatmap $C \in \mathbb{R}^{1, H / 4 \times W / 4}$. The instances in the frame are detected by taking the positions of the local maxima in $C$. This operation gives a set of $K$ positions, where $K$ is the number of detected instances and can be different for each frame. Afterwards, $K$ feature vectors - one for each detected instance - are generated by sampling each level of the backbone feature maps $F_i$ at the detected positions and concatenating the different levels. The resulting $K$ vectors are then reduced to 256-dimension and their positional encoding is added. This set of $K$ vectors is used together with the $N$ queries to generate an attention map using a single FFN layer, inner product, and softmax similarly to \cite{NIPS2017_3f5ee243}. The produced attention map values are used as mixing weights for the queries to output a set of new FSI queries.

The newly produced queries can then be added to the fixed ones obtaining a set of $N + K$ queries. In most applications, $K << N$; hence, the only significant computation overhead is given by the instance detector used. However, trying multiple detectors and evaluating their relative performance is outside the scope of our work. We observe that the new queries can significantly improve performance while also providing robustness with respect to the number of queries $N$. Additionally, the simplicity of the structure allows it to be applied to most query-based instance segmentation methods with mere adaptations of the implementation details.

\subsection{Training and Inference}\label{sec:train_inf}

SA-VIS base network (explained in Sec. \ref{sec:inst_arch}) is trained similarly to \cite{IDOL} using 4 different loss terms, namely $\mathcal{L}_{cls}$, $\mathcal{L}_{box}$, $\mathcal{L}_{mask}$ and $\mathcal{L}_{embed}$. We refer to \cite{IDOL} for more details regarding the meaning and computation of the loss terms. For clarity, we highlight that we sample pairs of annotated frames from a video and call them the key and reference frame during the training. The key frame is used to compute the instance segmentation losses, specifically $\mathcal{L}_{cls}$, $\mathcal{L}_{box}$, $\mathcal{L}_{mask}$. The reference frame, instead, is employed only as a source of contrastive embeddings and is used together with the key frame to compute $\mathcal{L}_{embed}$. 

We now focus on the training and inference schemes specific to the proposed method. During training, an additional loss is required to supervise the detector that is used to generate FSI queries. We call it $\mathcal{L}_{center}$ as the used detector predicts Gaussian heatmaps on the instance centers. $\mathcal{L}_{center}$ is the mean-squared error between the ground truth and the predicted heatmap. Similarly to the other instance loss terms, $\mathcal{L}_{center}$ is only applied to the key frame. Given all the components, the resulting total loss term is the following:
\begin{align}
    \begin{split}
        \label{eq:loss}
        \mathcal{L} = &\mathcal{L}_{cls} + \lambda_{box}\mathcal{L}_{box} + \lambda_{mask}\mathcal{L}_{mask} \\
        &+ \lambda_{embed}\mathcal{L}_{embed} + \lambda_{center}\mathcal{L}_{center}
    \end{split}
\end{align}
Here the weights parameters are set to $\lambda_{box} = 2.0$, $\lambda_{mask} = 2.0$, $\lambda_{embed} = 1.0$, $\lambda_{center} = 100.0$.

PFP, instead, does not require any custom loss in order to be trained. At each training iteration, in addition to the key-reference pair of annotated frames a set of $T$ other past frames (not necessarily annotated) are loaded. As explained in Section \ref{sec:past}, the $20$ past frames are used to generate the vectors $H_j$, which compose the FQ. The $H_j$ vectors are then included in the inference on the key and reference frame. We note that the FQ used when processing the reference frame includes also the vectors $H_{key}$. To reduce the memory and time complexity of training the model, no gradient is retained from the backbone used to generate the $H_j$ features and the transformer encoder-decoder is not run on them. As a consequence, the addition of PFP and FSI queries only add less than $10\%$ training time overhead. This stratagem also allows us to train the model without the need to jointly supervise the prediction of many frames (different to what is done in \cite{ying2023ctvis}). Thus SA-VIS can be trained to effectively include PFP with only $2$ NVIDIA A100 GPUs.

During inference, we follow the approach proposed by IDOL~\cite{IDOL}. The video frames are computed sequentially and instances are matched across frames based on their bi-softmax similarity \cite{fischer2023qdtrack} (See \cite{IDOL} for more details). On top of the base inference scheme, we build the FQ progressively as the frames are processed and use them for the following ones. In practice, the FQ works as a FIFO buffer, by adding new vectors $H_j$ and removing the vectors $H_{j - T}$. Older feature vectors are removed only after the first $T$ frames, during which the FQ is filled. When executing inference for each time step besides the first one, the vectors in the FQ are used in the network as described in Section \ref{sec:past}.

%% file: tables/ytvis192122.tex
\begin{table}[!t]
\caption{Comparisons on \textbf{YouTube-VIS 2019, 2021, and 2022 validation} sets.
The results are reported separately when using \textit{ResNet50} and \textit{Swin-L} backbone.
For each group, we \textbf{bold} the best values in every metric between the \emph{pairs train} methods. The best \emph{video train} method is \underline{underlined}. We note that SA-VIS not only strongly outperforms every other \emph{pairs train} by a large margin, but also significantly reduces the gap to \emph{video train}. We also highlight the very small gap in performance between SA-VIS and SA-VIS$_5$ which shows the effectiveness of the method in using sparsely annotated data.}
\label{tab:ytvis2019_2021_2022}
\centering\resizebox{0.95\linewidth}{!}
{ 
\begin{tabular}{@{}c|c|ccccc|ccccc|ccccc@{}}
\toprule
\multicolumn{2}{c|}{\multirow{2}{*}{\textbf{Method}}} & \multicolumn{5}{c|}{\textbf{YouTube-VIS 2019}} & \multicolumn{5}{c|}{\textbf{YouTube-VIS 2021}} & \multicolumn{5}{c}{\textbf{YouTube-VIS 2022$_{L}$}}\\
\multicolumn{2}{l|}{} & AP & AP$_{50}$ & AP$_{75}$ & AR$_1$ & AR$_{10}$ & AP & AP$_{50}$ & AP$_{75}$ & AR$_1$ & AR$_{10}$ & AP & AP$_{50}$ & AP$_{75}$ & AR$_1$ & AR$_{10}$ \\
    \midrule 
    \midrule
    \rowcolor{mygray} \cellcolor{white} & VITA~\cite{VITA}
    & 49.8  & 72.6  & 54.5  & 49.4  & 61.0
    & 45.7  & 67.4  & 49.5  & 40.9  & 53.6
    & 32.6  & 53.9  & 39.3  & 30.3  & 42.6 \\
    \rowcolor{mygray} \cellcolor{white} & GenVIS~\cite{GenVIS}
    & 50.0  & 71.5  & 54.6  & 49.5  & 59.7
    & 47.1  & 67.5  & 51.5  & 41.6  & 54.7
    & 37.5  & \underline{61.6}  & 41.5  & 32.6  & 42.2 \\
    \rowcolor{mygray} \cellcolor{white} & GRAtt-VIS~\cite{hannan2023gratt}
    & 50.4  & 70.7  & 55.2  & 48.4  & 58.7
    & 48.9  & 69.2  & 53.1  & 41.8  & 56.0
    & \underline{40.8}  & 60.1  & \underline{45.9}  & \underline{35.7}  & \underline{46.9} \\
    \rowcolor{mygray} \cellcolor{white} & LBVQ+SAM~\cite{10418101}
    & 52.5  & 74.8  & 57.8  & 50.1  & 60.0
    & 44.9  & 67.4  & 46.0  & 41.6  & 52.3
    & -  & -  & -  & -  & - \\
    \rowcolor{mygray} \cellcolor{white} & CTVIS~\cite{ying2023ctvis}
    & \underline{55.1}  & \underline{78.2}  & \underline{59.1}  & \underline{51.9}  & \underline{63.2}
    & \underline{50.1}  & \underline{73.7}  & \underline{54.7}  & \underline{41.8}  & \underline{59.5}
    & -     & -     & -     & -     & -    \\
    & MinVIS~\cite{huang2022minvis}
    & 47.4  & 69.0  & 52.1  & 45.7  & 55.7
    & 44.2  & 66.0  & 48.1  & 39.2  & 51.7
    & 23.2  & 47.9  & 19.3  & 20.2  & 28.0 \\
    & IDOL~\cite{IDOL}
    & 49.5  & 74.0  & 52.9  & 47.7  & 58.7
    & 43.9  & 68.0  & 49.6  & 38.0  & 50.9
    & -     & -     & -     & -     & -    \\
    & \textbf{SA-VIS (Ours)}
    & \textbf{51.2}  & \textbf{75.5}  & \textbf{54.8}  & \textbf{49.1}  & \textbf{60.4}
    & \textbf{48.5}  & \textbf{72.4}  & \textbf{52.3}  & \textbf{41.7}  & \textbf{57.2}
    & \textbf{38.6}  & \textbf{64.3}  & \textbf{42.7}  & \textbf{33.3}  & \textbf{44.5} \\
    \multirow{-9}{*}{\rotatebox{90}{\textit{ResNet50}}} & \textbf{SA-VIS$_5$ (Ours)}
    & 50.8  & 74.7  & 54.4  & 48.7  & 59.6
    & 48.1  & 71.8  & 52.0  & 41.1  & 56.4
    & 38.3  & 63.7  & 42.4  & 32.8  & 43.9 \\
    \midrule 
    \midrule
    \rowcolor{mygray} \cellcolor{white} & VITA~\cite{VITA}
    & 63.0  & 86.9  & 67.9  & 56.3  & 68.1
    & 57.5  & 80.6  & 61.0  & 47.7  & 62.6
    & 41.1  & 63.0  & 44.0  & 39.3  & 44.3 \\
    \rowcolor{mygray} \cellcolor{white} & GenVIS~\cite{GenVIS}
    & 64.0  & 84.9  & 68.3  & 56.1  & 69.4
    & 59.6  & 80.9  & 65.8  & 48.7  & 65.0
    & \underline{44.3}  & \underline{69.9}  & \underline{44.9}  & \underline{39.9}  & \underline{48.4} \\
    \rowcolor{mygray} \cellcolor{white} & CTVIS~\cite{ying2023ctvis}
    & \underline{65.6}  & \underline{87.7}  & \underline{72.2}  & \underline{56.5}  & \underline{70.4}
    & \underline{61.2}  & \underline{84.0}  & \underline{68.8}  & \underline{48.0}  & \underline{65.8}
    & -     & -     & -     & -     & -    \\
    & MinVIS~\cite{huang2022minvis}
    & 61.6  & 83.3  & 68.6  & 54.8  & 66.6
    & 55.3  & 76.6  & 62.0  & 45.9  & 60.8
    & 33.1  & 54.8  & 33.7  & 29.5  & 36.6 \\
    & IDOL~\cite{IDOL}
    & 64.3  & 87.5  & 71.0  & 55.5  & 69.1
    & 56.1  & 80.8  & 63.5  & 45.0  & 60.1
    & -     & -     & -     & -     & -    \\
    & \textbf{SA-VIS (Ours)}
    & \textbf{64.6}  & \textbf{88.5}  & \textbf{72.1}  & \textbf{56.0}  & \textbf{70.0}
    & \textbf{60.6}  & \textbf{83.7}  & \textbf{68.5}  & \textbf{47.8}  & \textbf{65.1}
    & \textbf{45.1}  & \textbf{71.3}  & \textbf{46.0}  & \textbf{40.8}  & \textbf{50.2} \\
    \multirow{-7}{*}{\rotatebox{90}{\textit{Swin-L}}} & \textbf{SA-VIS$_5$ (Ours)}
    & 64.0  & 87.6  & 71.6  & 55.4  & 69.4
    & 60.1  & 82.9  & 68.1  & 47.2  & 64.5
    & 44.6  & 70.5  & 45.4  & 40.3  & 49.6 \\
\bottomrule
\end{tabular}
} 
\end{table}

%% file: tables/ovis.tex
\begin{table}[!ht]
\centering
\parbox{.48\linewidth}{
\caption{Comparisons on \textbf{OVIS validation} set.
The results are grouped as \emph{video train} if they are trained on long video sequences or \emph{pairs train} if they only need couples of annotated images.
For each group, we \textbf{bold} the best values in every metric.}
\label{tab:ovis}
\centering\resizebox{0.99\linewidth}{!}
{ 
\begin{tabular}{@{}c|lc|ccccc@{}}
\toprule
\multicolumn{3}{c|}{\multirow{2}{*}{\textbf{Method}}} & \multicolumn{5}{c}{\textbf{OVIS}}\\
\multicolumn{3}{l|}{} & AP & AP$_{50}$ & AP$_{75}$ & AR$_1$ & AR$_{10}$ \\
    \midrule
    \midrule
    \multirow{9}{*}{\rotatebox{90}{\emph{video train}}}
    & \multicolumn{2}{l|}{IFC~\cite{hwang2021ifc}}
    & 13.1  & 27.8  & 11.6  &  9.4  & 23.9 \\
    & \multicolumn{2}{l|}{Mask2Former-VIS~\cite{Cheng2021Mask2FormerFV}}
    & 17.3  & 37.3  & 15.1  & 10.5  & 23.5  \\
    & \multicolumn{2}{l|}{TeViT~\cite{yang2022temporally}}
    & 17.4  & 34.9  & 15.0  & 11.2  & 21.8 \\
    & \multicolumn{2}{l|}{SeqFormer~\cite{wu2021seqformer}}
    & 15.1  & 31.9  & 13.8  & 10.4  & 27.1 \\   
    & \multicolumn{2}{l|}{VITA~\cite{VITA}}
    & 19.6  & 41.2  & 17.4  & 11.7  & 26.0 \\
    & \multicolumn{2}{l|}{GenVIS~\cite{GenVIS}}
    & 35.8  & \textbf{60.8}  & 36.2  & 16.3  & 39.6 \\
    & \multicolumn{2}{l|}{GRAtt-VIS~\cite{hannan2023gratt}}
    & \textbf{36.2}  & \textbf{60.8}  & \textbf{36.8}  & \textbf{16.8}  & 40.0 \\
    & \multicolumn{2}{l|}{LBVQ+SAM~\cite{10418101}}
    & 22.3  & 45.5  & 18.8  & 12.4  & 27.7 \\
    & \multicolumn{2}{l|}{CTVIS~\cite{ying2023ctvis}}
    & 35.5  & \textbf{60.8}  & 34.9  & 16.1  & \textbf{41.9} \\
    \midrule
    \midrule
    \multirow{7}{*}{\rotatebox{90}{\emph{pairs train}}}
    & \multicolumn{2}{l|}{CrossVIS~\cite{Yang_2021_ICCV}}
    & 14.9  & 32.7  & 12.1  & 10.3  & 19.8 \\
    & \multicolumn{2}{l|}{VISOLO~\cite{han2022visolo}}
    & 15.3  & 31.0  & 13.8  & 11.1  & 21.7 \\
    & \multicolumn{2}{l|}{InstanceFromer~\cite{koner2022instanceformer}}
    & 20.0  & 40.7  & 18.1  & 12.0  & 27.1 \\
    & \multicolumn{2}{l|}{MinVIS~\cite{huang2022minvis}}
    & 25.0  & 45.5  & 24.0  & 13.9  & 29.7 \\
    & \multicolumn{2}{l|}{IDOL~\cite{IDOL}}
    & 30.2  & 51.3  & 30.0  & 15.0  & 37.5 \\
    & \multicolumn{2}{l|}{\textbf{SA-VIS (Ours)}}
    & \textbf{32.9}  & \textbf{56.2}  & \textbf{34.3}  & \textbf{16.4}  & \textbf{40.9} \\
    & \multicolumn{2}{l|}{\textbf{SA-VIS$_5$ (Ours)}}
    & 32.5  & 55.6  & 34.0  & 15.9  & 40.2 \\
    \bottomrule
    \end{tabular}
} 
}
\hfill
\parbox{.48\linewidth}{
\caption{\textbf{Past Frame Features:} Ablation of the different components of the PFP branch. The base model contains FSI queries and no PFP. By FM influence (FMI), we indicate that the model applies the transformer on the FM but only feeds the features vectors $H_j$ for the current frame $j$. Feed FM (FFM) uses the FM as described in Section \ref{sec:past} but only storing sparse frames in the FM (1 every 4 frames). Dense FM (DFM) stores 20 consecutive frames. In all the considered cases, we build the FM using the features vectors from 20 consecutive frames.}
\label{tab:ablation_past}
\centering\resizebox{0.95\linewidth}{!}
{
\begin{tabular}{ccc|ccccc}
\toprule
\multirow{2}{*}{FMI} & \multirow{2}{*}{FFM} & \multirow{2}{*}{DFM} & \multicolumn{5}{c}{\textbf{YouTube-VIS 2021}}\\
&  & 
& AP & AP$_{50}$ & AP$_{75}$ & AR$_{1}$ & AR$_{10}$ \\
\midrule
\midrule
&  & 
& 45.5  & 68.2  & 50.7  & 35.8  & 55.3 \\
\cmark &   & 
& 46.1  & 70.5  & 50.9  & 40.1  & 55.7 \\
\cmark & \cmark & 
& 46.9  & 70.9  & 51.6  & 40.9  & \textbf{55.8} \\
\cmark & \cmark & \cmark
& \textbf{47.9}  & \textbf{71.7}  & \textbf{51.9}  & \textbf{41.2}  & 55.4 \\
\bottomrule
\end{tabular}
}
}
\end{table}

%% file: sec/4_experiment.tex
\section{Experiment}
\label{sec:experiment}

\input{tables/ablation_network}

\paragraph{Datasets and Metrics:} SA-VIS is evaluated on the standard VIS datasets YouTube-VIS 2019/2021/2022 (YTVIS19/YTVIS21/YTVIS22) \cite{Yang_2019_ICCV} and Occluded VIS (OVIS) \cite{qi2021ovis}. Among these datasets, YTVIS22 and OVIS are particularly challenging as they include respectively very long videos and many occluded instances. For YTVIS22, we report the results on the subset of long videos (YTVIS22$_L$) which does not overlap with YTVIS21. For evaluation, we follow common practice \cite{GenVIS, IDOL, Yang_2019_ICCV, ying2023ctvis, hannan2023gratt, VITA, Cheng2021Mask2FormerFV} and evaluate SA-VIS in terms of average precision (AP) and average recall (AR). When reporting the results, we divide the methods depending on whether they require joint training on long video sequences (\emph{video train} with \colorbox{mygray}{grey background}) or they only need single/couples of annotated images (\emph{pairs train} with white background).

In order to prove the ability of our method in exploiting the presence of unlabeled frames, we evaluate SA-VIS in two different training conditions. One is the standard, where annotations for every frame are used to train the model. The other, identified as \textbf{SA-VIS$_5$} trains the model using only a subset of the available annotations; only the annotations of 1 every 5 frames in the video are used, while the other 4 are considered unlabeled and used only as past frames in the SA.

\paragraph{Implementation Details:} For the network hyperparameters we refer to the implementation of \cite{IDOL} using ResNet-50 \cite{He_2016_CVPR} and Swin-L \cite{liu2021swin} as backbone. The models are pretrained on COCO \cite{lin2014microsoft} and trained on YouTubeVIS and OVIS using respectively $2$ or $3$ NVIDIA A100 GPUs. We note that this number is much lower compared to comparable \emph{video train} methods that require 8 GPUs of the same type \cite{ying2023ctvis}. As augmentations, we use clip-level random crop and flip, with clip defined as the key-reference frame and the past frames. Similar to \cite{IDOL}, images from YouTubeVIS datasets are downsampled to have the shortest side in the range of 320 to 640 pixels with the longest side cropped at 768. Additionally, the images have $0.5$ probability of being randomly cropped to size 384 $\times$ 600. Images from OVIS are treated similarly with changed resizing sizes from 480 to 736. During inference, YouTubeVIS and OVIS images are resized to 480 and 720 respectively. Training runs for 12\,000 iterations with batch size of 12 for YouTubeVIS and 9 for OVIS, starting learning rate at 0.0001 and weight decay with factor 0.1 after 8\,000 iterations.

\subsection{Main Results}\label{sec:results}

\paragraph{YTVIS19, YTVIS21, and YTVIS22:} On these datasets, which are the most established in the VIS field, SA-VIS results on par with SOTA performance, while outperforming by over $4\%$ every other comparable \emph{pairs train} method. Most interestingly, Table \ref{tab:ytvis2019_2021_2022} shows that SA-VIS consistently outperforms IDOL \cite{IDOL}, a method with which SA-VIS shares the training and evaluation scheme as well as the network except for the proposed SA and FSI queries. The consistently better performance of SA-VIS across all metrics, demonstrate the effectiveness of our proposed contributions.

\paragraph{OVIS:} The challenging dataset presents a larger number of medium and heavily occluded objects. Table \ref{tab:ovis} shows that also in this challenging context, SA-VIS outperforms every other \emph{pairs train} approach by a significant margin and performs on par with the powerful \emph{video train} on several metrics.

\paragraph{SA-VIS$_5$:} Separately from dataset-specific observations, we highlight that our method is extremely effective in exploiting even the not-annotated frames. Across all datasets, almost no performance drop is visible when using the annotations of only $1/5$ of the video frames.

\input{tables/ablation_nq}
\input{tables/ablation_matching}

\subsection{Ablation Studies}\label{sec:ablations}

We conduct several experiments to verify the effectiveness of the proposed solutions. All the experiments are conducted on YTVIS21 validation set.

\paragraph{Label Sparsity:} We first highlight that, as shown in the main results, SA-VIS is extremely effective in preserving high accuracy even when only sparse frames are labeled. In this paragraph, we isolate better the effect of the use of sparse annotations both with respect to other methods and different sparsity levels. Tab. \ref{tab:ablation_sparse} shows that SA-VIS is significantly more effective compared to other methods in using non-annotated frames. It also shows how the performance remains high even at very high levels of sparsity. Additionally, we observe that in equal conditions of pairwise training and sparse annotations, SA-VIS also outperforms SOTA method CTVIS \cite{ying2023ctvis}.

\paragraph{Network Components:} We first ablate both our 2 main contributions, namely the SA and the FSI queries. Table \ref{tab:ablation_network} clearly shows the benefit of each of our network additions as both provide significant improvement over the baseline. Applying both improves the results even further across all metrics. Interestingly, FSI queries which do not significantly help improve accuracy in terms of AP$_{50}$, have a bigger impact when applied together with PFP. This suggests that the two contributions have a complementary effect. For further analysis on the two aspects, we refer to the next paragraphs.

\paragraph{Past Frames Awareness:} In this paragraph, we ablate the components of the PFP architecture and demonstrate that PFP is effective in improving instance matching across frames.
Table \ref{tab:ablation_past} shows that each component in the architecture plays a role in improving the results. It is important both to feed the full processed FM as well as dense FM to the transformer encoder for the best performance.

In order to prove the intuition that adding PFP improves instance matching across frames, we design an experiment in which the instance masks are provided by an oracle. We thus divide the training set into two subsets, one for training and one for validation with oracle masks. On the validation sub-set, we run inference normally and, during postprocessing, we substitute the predicted masks with the best matching ground truth ones. Table \ref{tab:ablation_matching} clearly shows that PFP strongly improves the network's ability to match instances correctly across frames compared to the baseline without PFP. This result strongly suggests that PFP significantly boosts the instance matching capability of the network.

\input{figs/fig_tex/qualitative}

\paragraph{FSI Queries:} Here, we first show that the improved performance of adding FSI Queries is not due to the higher network capacity. In Table \ref{tab:ablation_nq}, we observe the performance when changing the number of fixed queries $N$. We can see that performance of our method is very stable in the range between 100 and 400 fixed queries. This result proves that the performance improvement of the added FSI queries cannot be attributed to the number of added queries which is much smaller ($K = 5.4$). Additionally, adding FSI queries makes the method robust against the changing number of fixed queries.

We perform an experiment to evaluate the ability of FSI queries to capture better masks compared to the fixed queries. We divide the training set similarly to the PFP oracle experiment. We evaluate the instance prediction quality by considering the 10 best matching predicted masks. We observe that overall, for $82\%$ of the instances, predictions obtained by the FSI queries are among the $10$ best matching ones. The ratio is $96\%$ when excluding the instances that are not detected using center heatmap prediction. Additionally, in $47\%$ of the cases, the instances predicted through FSI queries are the top-$1$ best match to the ground truth mask. This implies that when the light weight detector finds an instance, FSI queries can generate an accurate mask. The results are even more significant when considering that we have $N = 300$ fixed queries and on average only $K = 5.4$ FSI queries.

\subsection{Qualitative Results}\label{sec:qualit}

In Figure \ref{fig:qualitative} we show the performance of our method on very challenging scenarios. The shown videos either display fast motion, low resolution images, similarly textured and cluttered objects and occlusion. In all the shown cases, our method generates accurate segmentation masks and successfully matches instances across frames.

%% file: tables/ablation_network.tex
\begin{table}[!t]
\centering
\parbox{.44\linewidth}{
\caption{\textbf{Network Components:} Ablation of the newly proposed network components. PFA indicates the past-frames awareness and FSIQ indicates the FSI queries.}
\label{tab:ablation_network}
\centering\resizebox{\linewidth}{!}
{
\begin{tabular}{cc|ccccc}
\toprule
\multirow{2}{*}{PFA} & \multirow{2}{*}{FSIQ} & \multicolumn{5}{c}{\textbf{YouTube-VIS 2021}}\\
&
& AP & AP$_{50}$ & AP$_{75}$ & AR$_{1}$ & AR$_{10}$ \\
\midrule
\midrule
& 
& 43.9  & 68.0  & 49.6  & 38.0  & 50.9 \\
\cmark &  
& 46.5  & 70.2  & 51.2  & 39.5  & 54.0 \\
 & \cmark
& 45.5  & 68.2  & 50.7  & 35.8  & 55.3 \\
\cmark & \cmark
& \textbf{48.5}  & \textbf{72.4}  & \textbf{52.3}  & \textbf{41.7}  & \textbf{57.2} \\
\bottomrule
\end{tabular}
} }
\hfill
\parbox{.52\linewidth}{
\caption{\textbf{Annotation Sparsity:} 
Comparison of the performance when using different annotation sparsity levels (SL). 1 means dense annotations; 2 means one frame every 2 is annotated \emph{etc}. Each method is trained using pairs couples as in \emph{pairs train}.}
\label{tab:ablation_sparse}
\centering\resizebox{0.85\linewidth}{!}
{
\begin{tabular}{c|c|ccccc}
\toprule
\multirow{2}{*}{\textbf{Method}} & \multirow{2}{*}{\textbf{SL}} & \multicolumn{5}{c}{\textbf{YouTube-VIS 2021}}\\
& 
& AP & AP$_{50}$ & AP$_{75}$ & AR$_{1}$ & AR$_{10}$ \\
\midrule
\midrule
\multirow{2}{*}{CTVIS\cite{ying2023ctvis}} & 1
& 48.3  & 71.4  & 52.4  & 41.8  & 57.5 \\
& 5
& 47.0  & 69.5  & 51.1  & 41.1  & 56.7 \\
\multirow{2}{*}{IDOL\cite{IDOL}} & 1
& 43.9  & 68.0  & 49.6  & 38.0  & 50.9 \\
& 5
& 42.8  & 66.7  & 48.7  & 37.4  & 50.0 \\\midrule
\multirow{4}{*}{\textbf{SA-VIS}} & 1
& 48.5  & 72.4  & 52.3  & 41.7  & 57.2 \\
& 2 
& 48.3  & 72.2  & 52.2  & 41.4  & 56.9 \\
& 5
& 48.1  & 71.8  & 52.0  & 41.1  & 56.4 \\
& 10
& 46.5  & 68.3  & 49.2  & 40.6  & 56.0 \\
\bottomrule
\end{tabular}
} }

\end{table}

%% file: tables/ablation_nq.tex
\begin{table}[!t]
\caption{\textbf{Number of Queries:} The effect of FSI queries is highlighted by changing the number of fixed queries. The number of queries $N = 300$ used in the other experiments is underlined and the performance gap when changing $N$ is highlighted.}
\label{tab:ablation_nq}
\centering\resizebox{0.7\linewidth}{!}
{
\begin{tabular}{c|cc|cc}
\toprule
\multirow{2}{*}{\# of fixed queries} & \multicolumn{2}{c|}{w/ FSI queries} & \multicolumn{2}{c}{w/o FSI queries}\\
& AP & AR$_{10}$ & AP & AR$_{10}$ \\
\midrule
\midrule
50  & 47.6 \textcolor{red}{(-0.9)}  & 55.9 \textcolor{red}{(-1.3)}  & 43.4 \textcolor{red}{(-3.1)}  & 51.2 \textcolor{red}{(-4.2)} \\
100 & 48.7 \textcolor{red}{(+0.2)}  & 57.3 \textcolor{red}{(+0.1)}  & 44.9 \textcolor{red}{(-1.6)}  & 53.5 \textcolor{red}{(-0.5)} \\
200 & 48.4 \textcolor{red}{(-0.1)}  & 57.0 \textcolor{red}{(-0.2)}  & 45.8 \textcolor{red}{(-0.7)}  & 53.8 \textcolor{red}{(-0.2)} \\
\underline{300} & \underline{48.5}  & \underline{57.2}  & \underline{46.5}  & \underline{54.0} \\
400 & 48.2 \textcolor{red}{(-0.3)}  & 57.5 \textcolor{red}{(+0.3)}  & 46.2 \textcolor{red}{(-0.3)}  & 54.5 \textcolor{red}{(+0.5)} \\
\bottomrule
\end{tabular}
}
\end{table}

%% file: tables/ablation_matching.tex
\begin{table}[!h]
\parbox{.48\linewidth}{
\caption{\textbf{Effect of PFP on Matching Accuracy:} The effect of PFP on matching instances across frames is isolated by providing oracle mask predictions.}
\label{tab:ablation_matching}
\centering\resizebox{0.85\linewidth}{!}
{
\begin{tabular}{c|ccccc}
\toprule
\multirow{2}{*}{PFP} & \multicolumn{5}{c}{\textbf{YouTube-VIS 2021}}\\
& AP & AP$_{50}$ & AP$_{75}$ & AR$_{1}$ & AR$_{10}$ \\
\midrule
\midrule
        & 58.0  & 82.2  & 68.1  & 52.5  & 60.3 \\
\cmark  & \textbf{65.0}  & \textbf{87.4}  & \textbf{74.2}  & \textbf{56.1}  & \textbf{65.8} \\
\bottomrule
\end{tabular}
}}
\hfill
\parbox{.48\linewidth}{
\caption{\textbf{Performance Evaluation}: The additions of SA-VIS only add marginal complexity to the original IDOL architecture.}
\label{tab:perf_reb}
\centering
\resizebox{0.9\linewidth}{!}
{ 
\begin{tabular}{c|ccc}
\toprule
\multirow{2}{*}{\textit{Method}} & \multicolumn{3}{c}{\textit{Metrics}}\\
& MParams & GFLOPs & FPS\\
\midrule
\midrule
IDOL & 48  & 100  & 30 \\
\textbf{SA-VIS (Ours)} & 51 (+6\%)  & 109 (+9\%)  & 28 (-7\%) \\
\bottomrule
\end{tabular}
}} 
\end{table}

%% file: figs/fig_tex/qualitative.tex
\begin{figure}
\centering
\includegraphics[width=0.27\linewidth]{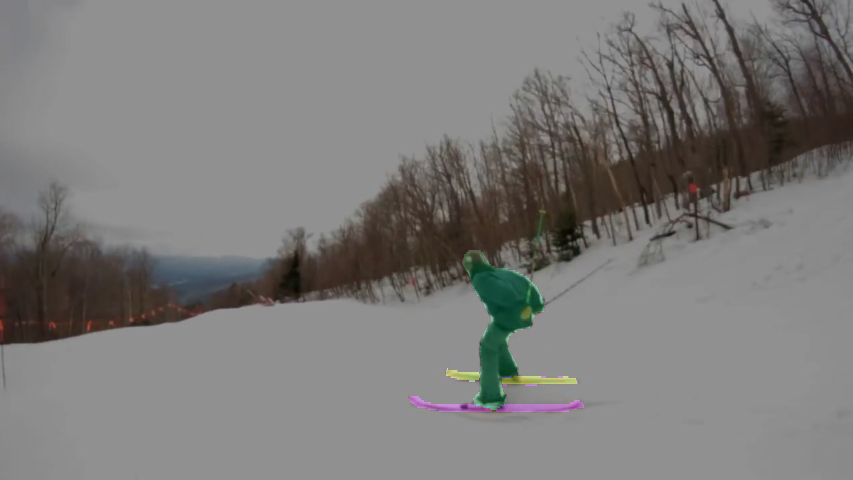}
\hspace{0.015\linewidth}
\includegraphics[width=0.27\linewidth]{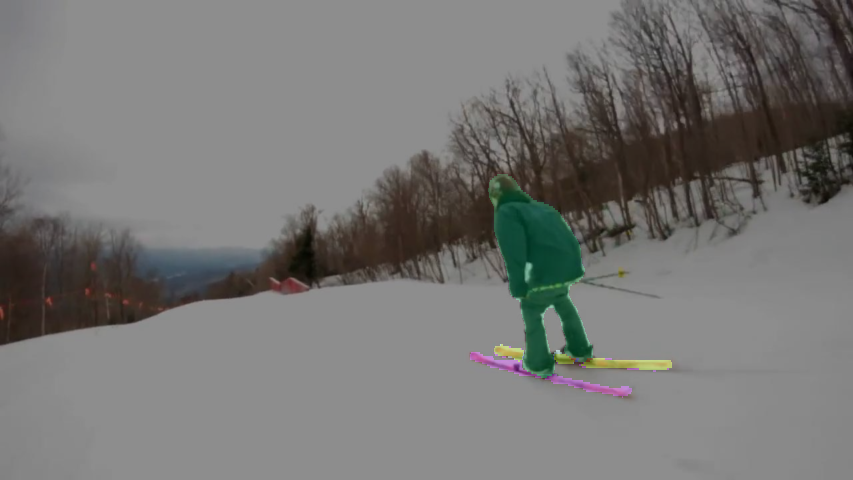}
\hspace{0.015\linewidth}
\includegraphics[width=0.27\linewidth]{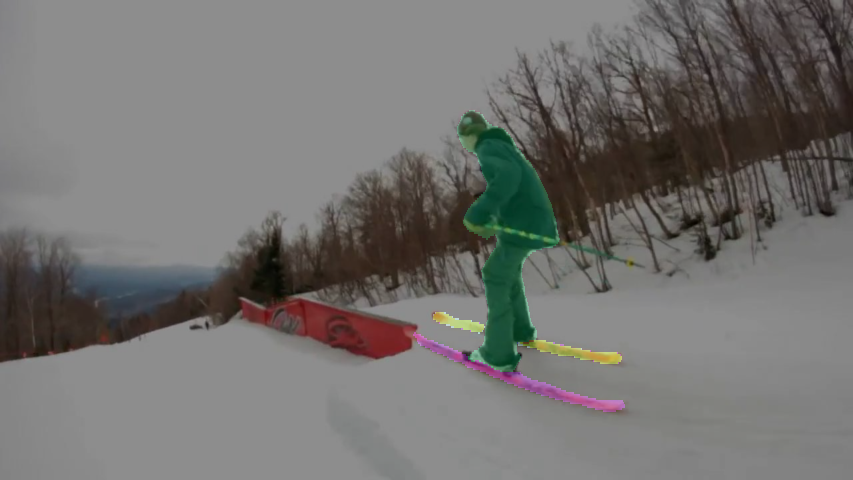} \\
\vspace{5pt}
\includegraphics[width=0.27\linewidth]{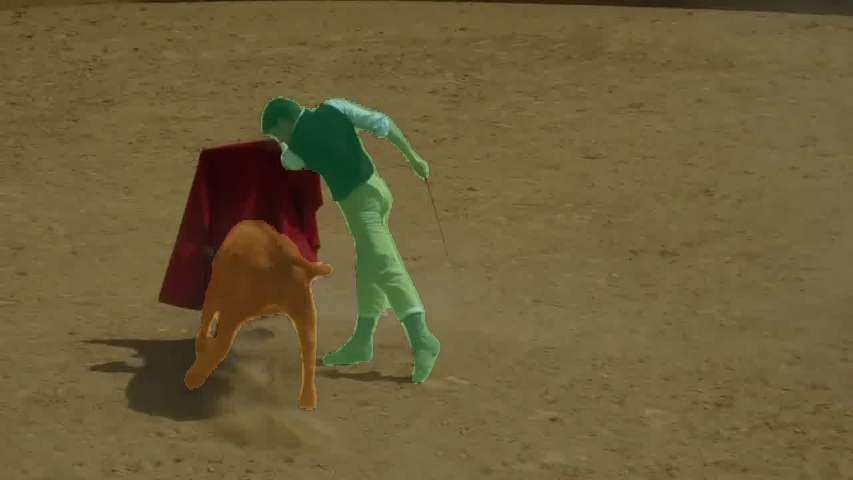}
\hspace{0.015\linewidth}
\includegraphics[width=0.27\linewidth]{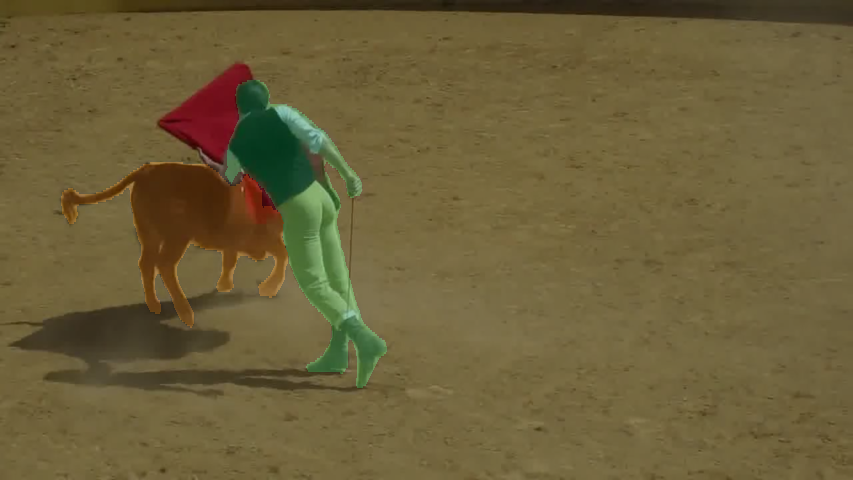}
\hspace{0.015\linewidth}
\includegraphics[width=0.27\linewidth]{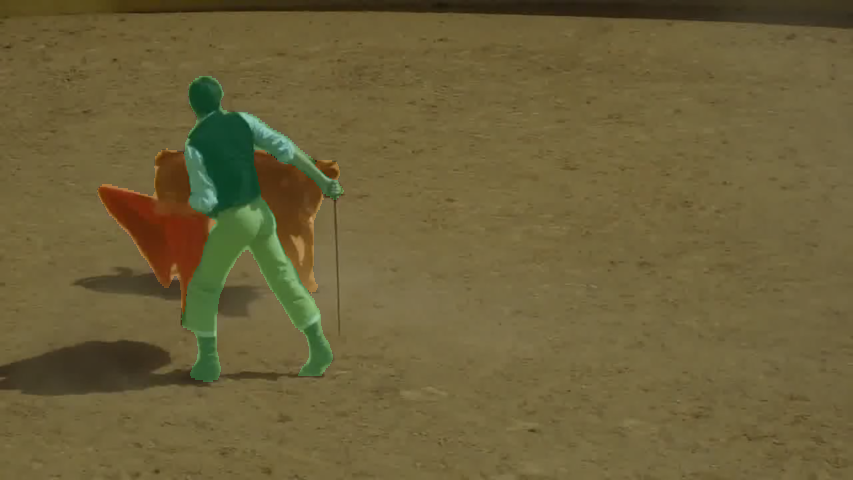} \\
\vspace{5pt}
\includegraphics[width=0.27\linewidth]{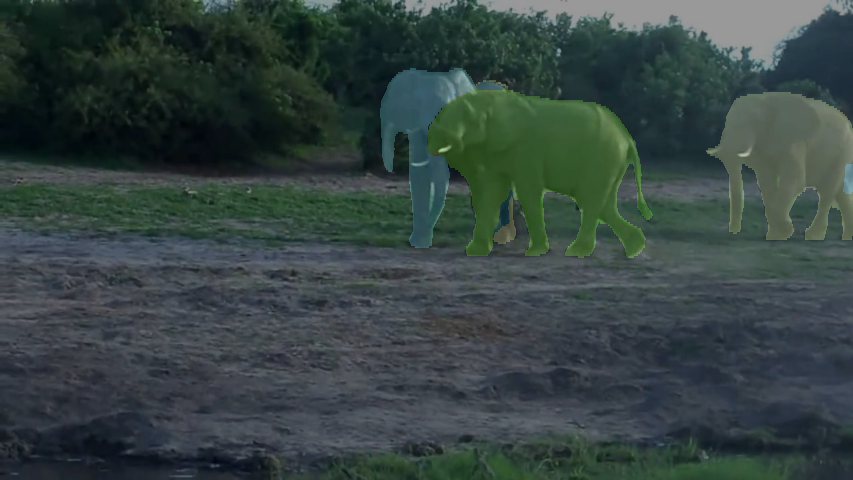}
\hspace{0.015\linewidth}
\includegraphics[width=0.27\linewidth]{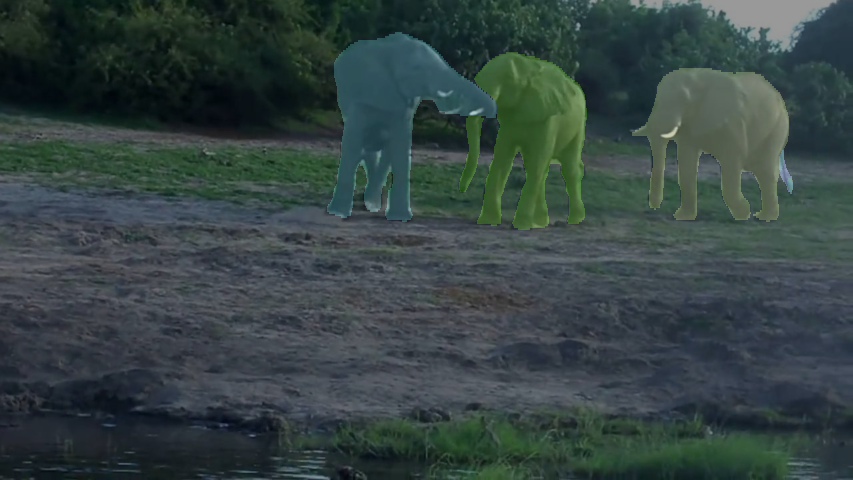}
\hspace{0.015\linewidth}
\includegraphics[width=0.27\linewidth]{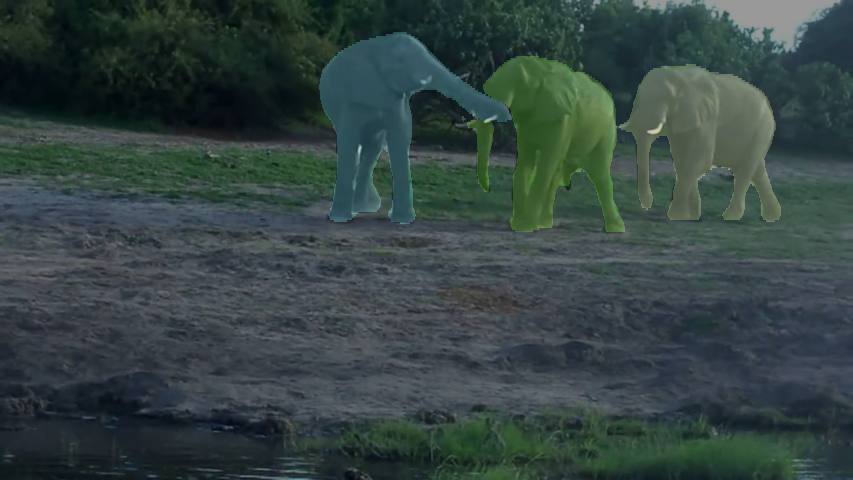} \\
\vspace{5pt}
\includegraphics[width=0.27\linewidth]{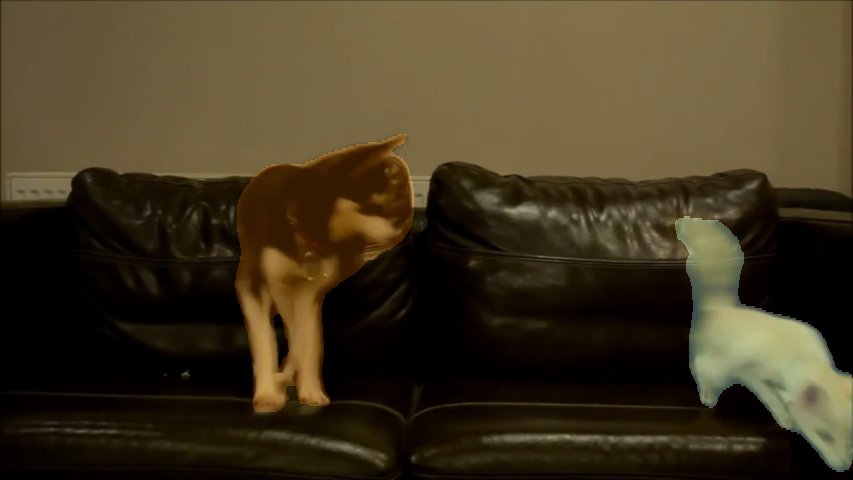}
\hspace{0.015\linewidth}
\includegraphics[width=0.27\linewidth]{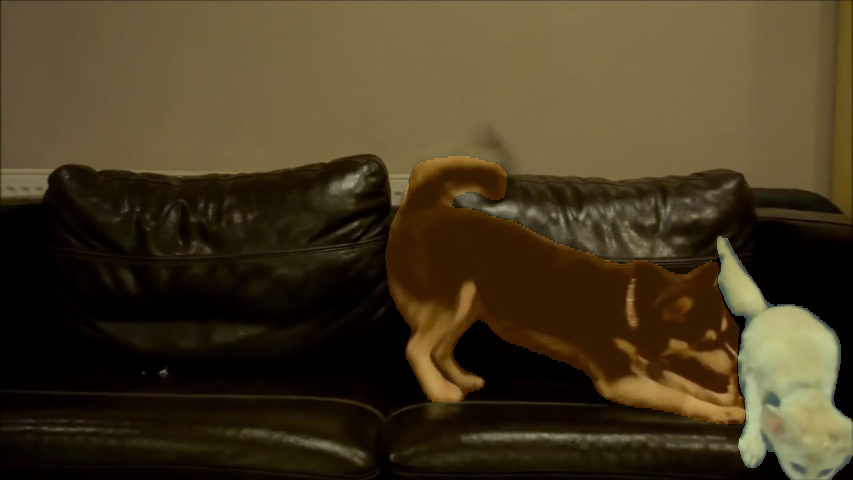}
\hspace{0.015\linewidth}
\includegraphics[width=0.27\linewidth]{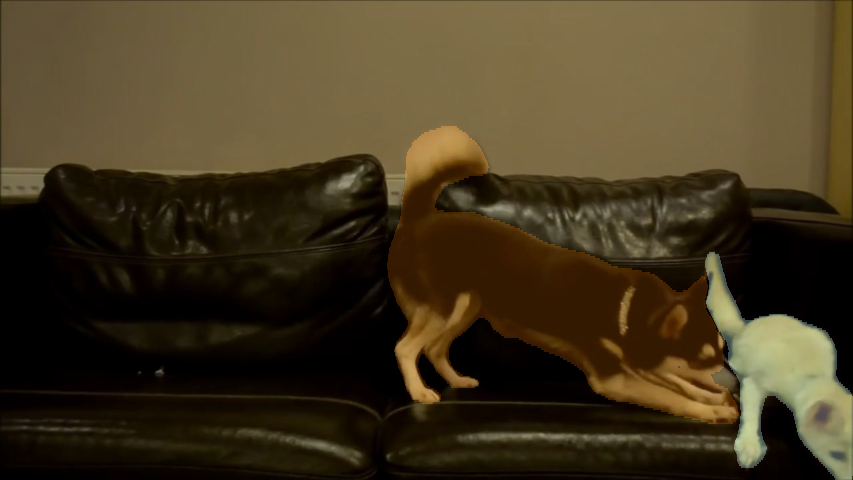} 
\caption{\textbf{Qualitative results:} Visualization of SA-VIS on a set of challenging scenes.}
\label{fig:qualitative}
\end{figure}

%% file: sec/5_limitations.tex
\section{Limitations and Future Works}
\label{sec:limitations}

The proposed method achieves state-of-the-art results and performs particularly well when sparse frame annotations are available. These results together with the ablations suggest that the proposed PFP and FSI queries are two effective ways to improve VIS performance. However, the simple method can be improved in several ways. For example, long videos in which similar scenes recur at distant time steps could benefit from varying and adaptive selection of the past frames to attend to.
Additionally, the use of high number of fixed queries may be unnecessary if there exists a strong instance detector in the FSI queries branch. This may be achieved with a detector that has high recall at the cost of false positives with the aim of ensuring all instances have corresponding FSI queries. Such an approach is interesting to explore as it would reduce the required computation while potentially simplifying the matching task.
A further possible future direction could focus on applying SA-VIS to very large sparsely annotated datasets to maximize the benefit of the proposed method.

%% file: sec/6_conclusion.tex
\section{Conclusion}
\label{sec:conclusion}

In this paper, we propose SA-VIS, an online VIS method, designed with the ability to train with sparse frame annotations. By adding attention to a set of features from the past frames, the model improves instance matching across frames. In parallel, using a set of FSI queries allows the model to predict more accurate masks. The proposed additions significantly improve performance in conditions of sparse annotations, while keeping the training simple and lightweight. We present results and ablations that suggest that these are viable approaches to achieve high accuracy in VIS. Despite having low complexity in design, they provide strong results for future research on compute and data efficient training of VIS.

%% file: main.bib
@String(CVPR= {IEEE Conf. Comput. Vis. Pattern Recog.})

@String(ICCV= {Int. Conf. Comput. Vis.})

@String(ECCV= {Eur. Conf. Comput. Vis.})

@String(AAAI = {AAAI})

@String(CVPR  = {CVPR})

@String(ICCV  = {ICCV})

@String(ECCV  = {ECCV})

@inproceedings{Yang_2019_ICCV,
author = {Yang, Linjie and Fan, Yuchen and Xu, Ning},
title = {Video Instance Segmentation},
booktitle = {Proceedings of the IEEE/CVF International Conference on Computer Vision (ICCV)},
month = {October},
year = {2019}
}

@inproceedings{liu2021swin,
  title={Swin transformer: Hierarchical vision transformer using shifted windows},
  author={Liu, Ze and Lin, Yutong and Cao, Yue and Hu, Han and Wei, Yixuan and Zhang, Zheng and Lin, Stephen and Guo, Baining},
  booktitle={Proceedings of the IEEE/CVF international conference on computer vision},
  pages={10012--10022},
  year={2021}
}

@inproceedings{qi2021ovis,
title={Occluded Video Instance Segmentation: Dataset and {ICCV} 2021 Challenge},
author={Jiyang Qi and Yan Gao and Yao Hu and Xinggang Wang and Xiaoyu Liu and Xiang Bai and Serge Belongie and Alan Yuille and Philip Torr and Song Bai},
booktitle={Thirty-fifth Conference on Neural Information Processing Systems Datasets and Benchmarks Track (Round 2)},
year={2021},
url={https://openreview.net/forum?id=IfzTefIU_3j}
}

@inproceedings{IDOL,
  title={In Defense of Online Models for Video Instance Segmentation},
  author={Wu, Junfeng and Liu, Qihao and Jiang, Yi and Bai, Song and Yuille, Alan and Bai, Xiang},
  booktitle={ECCV},
  year={2022},
}

@article{hannan2023gratt,
  title={GRAtt-VIS: Gated Residual Attention for Auto Rectifying Video Instance Segmentation},
  author={Hannan, Tanveer and Koner, Rajat and Bernhard, Maximilian and Shit, Suprosanna and Menze, Bjoern and Tresp, Volker and Schubert, Matthias and Seidl, Thomas},
  journal={arXiv preprint arXiv:2305.17096},
  year={2023}
}

@misc{ying2023ctvis,
      title={{CTVIS}: {C}onsistent {T}raining for {O}nline {V}ideo {I}nstance {S}egmentation}, 
      author={Kaining Ying and Qing Zhong and Weian Mao and Zhenhua Wang and Hao Chen and Lin Yuanbo Wu and Yifan Liu and Chengxiang Fan and Yunzhi Zhuge and Chunhua Shen},
      year={2023},
      eprint={2307.12616},
      archivePrefix={arXiv},
      primaryClass={cs.CV}
}

@InProceedings{Yang_2021_ICCV,
    author    = {Yang, Shusheng and Fang, Yuxin and Wang, Xinggang and Li, Yu and Fang, Chen and Shan, Ying and Feng, Bin and Liu, Wenyu},
    title     = {Crossover Learning for Fast Online Video Instance Segmentation},
    booktitle = {Proceedings of the IEEE/CVF International Conference on Computer Vision (ICCV)},
    month     = {October},
    year      = {2021},
    pages     = {8043-8052}
}

@inproceedings{huang2022minvis,
  title={MinVIS: A Minimal Video Instance Segmentation Framework without Video-based Training},
  author={De-An Huang and Zhiding Yu and Anima Anandkumar},
  journal={NeurIPS},
  year={2022}
}

@inproceedings{maskfreevis,
    author={Ke, Lei and Danelljan, Martin and Ding, Henghui and Tai, Yu-Wing and Tang, Chi-Keung and Yu, Fisher},
    title={Mask-Free Video Instance Segmentation},
    booktitle = {CVPR},
    year = {2023}
}

@article{Cheng2021Mask2FormerFV,
  title={Mask2Former for Video Instance Segmentation},
  author={Bowen Cheng and Anwesa Choudhuri and Ishan Misra and Alexander Kirillov and Rohit Girdhar and Alexander G. Schwing},
  journal={ArXiv},
  year={2021},
  volume={abs/2112.10764},
  url={https://api.semanticscholar.org/CorpusID:245335013}
}

@inproceedings{VITA,
  title={VITA: Video Instance Segmentation via Object Token Association},
  author={Heo, Miran and Hwang, Sukjun and Oh, Seoung Wug and Lee, Joon-Young and Kim, Seon Joo},
  booktitle={Advances in Neural Information Processing Systems},
  year={2022}
}

@inproceedings{Jiang2023STC,
author = {Jiang, Zhengkai and Gu, Zhangxuan and Peng, Jinlong and Zhou, Hang and Liu, Liang and Wang, Yabiao and Tai, Ying and Wang, Chengjie and Zhang, Liqing},
title = {STC: Spatio-Temporal Contrastive Learning For Video Instance Segmentation},
year = {2023},
booktitle = {Computer Vision – ECCV 2022 Workshops: Tel Aviv, Israel, October 23–27, 2022, Proceedings, Part IV},
location = {Tel Aviv, Israel}
}

@inproceedings{wang2020end,
  title={End-to-End Video Instance Segmentation with Transformers},
  author={Wang, Yuqing and Xu, Zhaoliang and Wang, Xinlong and Shen, Chunhua and Cheng, Baoshan and Shen, Hao and Xia, Huaxia},
  booktitle =  {Proc. IEEE Conf. Computer Vision and Pattern Recognition (CVPR)},
  year={2021}
}

@article{wu2021seqformer,
      title={SeqFormer: a Frustratingly Simple Model for Video Instance Segmentation}, 
      author={Junfeng Wu and Yi Jiang and Wenqing Zhang and Xiang Bai and Song Bai},
      journal={arXiv preprint arXiv:2112.08275},
      year={2021},
}

@inproceedings{GenVIS,
  title={A Generalized Framework for Video Instance Segmentation},
  author={Heo, Miran and Hwang, Sukjun and Hyun, Jeongseok and Kim, Hanjung and Oh, Seoung Wug and Lee, Joon-Young and Kim, Seon Joo},
  booktitle={CVPR},
  year={2023}
}

@InProceedings{Choudhuri_2023_CVPR,
    author    = {Choudhuri, Anwesa and Chowdhary, Girish and Schwing, Alexander G.},
    title     = {Context-Aware Relative Object Queries To Unify Video Instance and Panoptic Segmentation},
    booktitle = {Proceedings of the IEEE/CVF Conference on Computer Vision and Pattern Recognition (CVPR)},
    month     = {June},
    year      = {2023},
    pages     = {6377-6386}
}

@inproceedings{NIPS2017_3f5ee243,
 author = {Vaswani, Ashish and Shazeer, Noam and Parmar, Niki and Uszkoreit, Jakob and Jones, Llion and Gomez, Aidan N and Kaiser, \L ukasz and Polosukhin, Illia},
 booktitle = {Advances in Neural Information Processing Systems},
 editor = {I. Guyon and U. Von Luxburg and S. Bengio and H. Wallach and R. Fergus and S. Vishwanathan and R. Garnett},
 pages = {},
 publisher = {Curran Associates, Inc.},
 title = {Attention is All you Need},
 url = {https://proceedings.neurips.cc/paper_files/paper/2017/file/3f5ee243547dee91fbd053c1c4a845aa-Paper.pdf},
 volume = {30},
 year = {2017}
}

@inproceedings{
zhu2021deformable,
title={Deformable {\{}DETR{\}}: Deformable Transformers for End-to-End Object Detection},
author={Xizhou Zhu and Weijie Su and Lewei Lu and Bin Li and Xiaogang Wang and Jifeng Dai},
booktitle={International Conference on Learning Representations},
year={2021},
url={https://openreview.net/forum?id=gZ9hCDWe6ke}
}

@inproceedings{cheng2021mask2former,
  title={Masked-attention Mask Transformer for Universal Image Segmentation},
  author={Bowen Cheng and Ishan Misra and Alexander G. Schwing and Alexander Kirillov and Rohit Girdhar},
  journal={CVPR},
  year={2022}
}

@article{hwang2021ifc,
  title={Video instance segmentation using inter-frame communication transformers},
  author={Hwang, Sukjun and Heo, Miran and Oh, Seoung Wug and Kim, Seon Joo},
  journal={Advances in Neural Information Processing Systems},
  volume={34},
  pages={13352--13363},
  year={2021}
}

@inproceedings{carion2020end,
  title={End-to-end object detection with transformers},
  author={Carion, Nicolas and Massa, Francisco and Synnaeve, Gabriel and Usunier, Nicolas and Kirillov, Alexander and Zagoruyko, Sergey},
  booktitle={European conference on computer vision},
  pages={213--229},
  year={2020},
  organization={Springer}
}

@inproceedings{cheng2021maskformer,
  title={Per-Pixel Classification is Not All You Need for Semantic Segmentation},
  author={Bowen Cheng and Alexander G. Schwing and Alexander Kirillov},
  journal={NeurIPS},
  year={2021}
}

@InProceedings{He_2017_ICCV,
author = {He, Kaiming and Gkioxari, Georgia and Dollar, Piotr and Girshick, Ross},
title = {Mask R-CNN},
booktitle = {Proceedings of the IEEE International Conference on Computer Vision (ICCV)},
month = {Oct},
year = {2017}
}

@InProceedings{Voigtlaender_2019_CVPR,
author = {Voigtlaender, Paul and Krause, Michael and Osep, Aljosa and Luiten, Jonathon and Sekar, Berin Balachandar Gnana and Geiger, Andreas and Leibe, Bastian},
title = {MOTS: Multi-Object Tracking and Segmentation},
booktitle = {Proceedings of the IEEE/CVF Conference on Computer Vision and Pattern Recognition (CVPR)},
month = {June},
year = {2019}
}

@inproceedings{porzi2020learning,
  title={Learning multi-object tracking and segmentation from automatic annotations},
  author={Porzi, Lorenzo and Hofinger, Markus and Ruiz, Idoia and Serrat, Joan and Bulo, Samuel Rota and Kontschieder, Peter},
  booktitle={Proceedings of the IEEE/CVF Conference on Computer Vision and Pattern Recognition},
  pages={6846--6855},
  year={2020}
}

@inproceedings{lin2020video,
  title={Video instance segmentation tracking with a modified vae architecture},
  author={Lin, Chung-Ching and Hung, Ying and Feris, Rogerio and He, Linglin},
  booktitle={Proceedings of the IEEE/CVF Conference on Computer Vision and Pattern Recognition},
  pages={13147--13157},
  year={2020}
}

@inproceedings{cao2020sipmask,
  title={Sipmask: Spatial information preservation for fast image and video instance segmentation},
  author={Cao, Jiale and Anwer, Rao Muhammad and Cholakkal, Hisham and Khan, Fahad Shahbaz and Pang, Yanwei and Shao, Ling},
  booktitle={Computer Vision--ECCV 2020: 16th European Conference, Glasgow, UK, August 23--28, 2020, Proceedings, Part XIV 16},
  pages={1--18},
  year={2020},
  organization={Springer}
}

@inproceedings{fu2021compfeat,
  title={Compfeat: Comprehensive feature aggregation for video instance segmentation},
  author={Fu, Yang and Yang, Linjie and Liu, Ding and Huang, Thomas S and Shi, Humphrey},
  booktitle={Proceedings of the AAAI Conference on Artificial Intelligence},
  volume={35},
  number={2},
  pages={1361--1369},
  year={2021}
}

@inproceedings{wu2021track,
  title={Track to detect and segment: An online multi-object tracker},
  author={Wu, Jialian and Cao, Jiale and Song, Liangchen and Wang, Yu and Yang, Ming and Yuan, Junsong},
  booktitle={Proceedings of the IEEE/CVF conference on computer vision and pattern recognition},
  pages={12352--12361},
  year={2021}
}

@inproceedings{liu2021sg,
  title={Sg-net: Spatial granularity network for one-stage video instance segmentation},
  author={Liu, Dongfang and Cui, Yiming and Tan, Wenbo and Chen, Yingjie},
  booktitle={Proceedings of the IEEE/CVF Conference on Computer Vision and Pattern Recognition},
  pages={9816--9825},
  year={2021}
}

@article{koner2022instanceformer,
  title={InstanceFormer: An Online Video Instance Segmentation Framework},
  author={Koner, Rajat and Hannan, Tanveer and Shit, Suprosanna and Sharifzadeh, Sahand and Schubert, Matthias and Seidl, Thomas and Tresp, Volker},
  journal={arXiv preprint arXiv:2208.10547},
  year={2022}
}

@article{zhan2022robust,
  title={Robust online video instance segmentation with track queries},
  author={Zhan, Zitong and McKee, Daniel and Lazebnik, Svetlana},
  journal={arXiv preprint arXiv:2211.09108},
  year={2022}
}

@InProceedings{Zhang_2023_ICCV,
    author    = {Zhang, Tao and Tian, Xingye and Wu, Yu and Ji, Shunping and Wang, Xuebo and Zhang, Yuan and Wan, Pengfei},
    title     = {DVIS: Decoupled Video Instance Segmentation Framework},
    booktitle = {Proceedings of the IEEE/CVF International Conference on Computer Vision (ICCV)},
    month     = {October},
    year      = {2023},
    pages     = {1282-1291}
}

@inproceedings{wu2022efficient,
  title={Efficient video instance segmentation via tracklet query and proposal},
  author={Wu, Jialian and Yarram, Sudhir and Liang, Hui and Lan, Tian and Yuan, Junsong and Eledath, Jayan and Medioni, Gerard},
  booktitle={Proceedings of the IEEE/CVF Conference on Computer Vision and Pattern Recognition},
  pages={959--968},
  year={2022}
}

@ARTICLE{10418101,

  author={Fang, Hao and Zhang, Tong and Zhou, Xiaofei and Zhang, Xinxin},

  journal={IEEE Transactions on Circuits and Systems for Video Technology}, 

  title={Learning Better Video Query with SAM for Video Instance Segmentation}, 

  year={2024},

  volume={},

  number={},

  pages={1-1},

  doi={10.1109/TCSVT.2024.3361076}}

@InProceedings{He_2016_CVPR,
author = {He, Kaiming and Zhang, Xiangyu and Ren, Shaoqing and Sun, Jian},
title = {Deep Residual Learning for Image Recognition},
booktitle = {Proceedings of the IEEE Conference on Computer Vision and Pattern Recognition (CVPR)},
month = {June},
year = {2016}
}

@inproceedings{tian2020conditional,
  title={Conditional convolutions for instance segmentation},
  author={Tian, Zhi and Shen, Chunhua and Chen, Hao},
  booktitle={Computer Vision--ECCV 2020: 16th European Conference, Glasgow, UK, August 23--28, 2020, Proceedings, Part I 16},
  pages={282--298},
  year={2020},
  organization={Springer}
}

@InProceedings{Cheng_2020_CVPR,
author = {Cheng, Bowen and Collins, Maxwell D. and Zhu, Yukun and Liu, Ting and Huang, Thomas S. and Adam, Hartwig and Chen, Liang-Chieh},
title = {Panoptic-DeepLab: A Simple, Strong, and Fast Baseline for Bottom-Up Panoptic Segmentation},
booktitle = {Proceedings of the IEEE/CVF Conference on Computer Vision and Pattern Recognition (CVPR)},
month = {June},
year = {2020}
}

@misc{fischer2023qdtrack,
      title={QDTrack: Quasi-Dense Similarity Learning for Appearance-Only Multiple Object Tracking}, 
      author={Tobias Fischer and Thomas E. Huang and Jiangmiao Pang and Linlu Qiu and Haofeng Chen and Trevor Darrell and Fisher Yu},
      year={2023},
      eprint={2210.06984},
      archivePrefix={arXiv},
      primaryClass={cs.CV}
}

@inproceedings{lin2014microsoft,
  title={Microsoft coco: Common objects in context},
  author={Lin, Tsung-Yi and Maire, Michael and Belongie, Serge and Hays, James and Perona, Pietro and Ramanan, Deva and Doll{\'a}r, Piotr and Zitnick, C Lawrence},
  booktitle={Computer Vision--ECCV 2014: 13th European Conference, Zurich, Switzerland, September 6-12, 2014, Proceedings, Part V 13},
  pages={740--755},
  year={2014},
  organization={Springer}
}

@inproceedings{han2022visolo,
  title={Visolo: Grid-based space-time aggregation for efficient online video instance segmentation},
  author={Han, Su Ho and Hwang, Sukjun and Oh, Seoung Wug and Park, Yeonchool and Kim, Hyunwoo and Kim, Min-Jung and Kim, Seon Joo},
  booktitle={Proceedings of the IEEE/CVF Conference on Computer Vision and Pattern Recognition},
  pages={2896--2905},
  year={2022}
}

@inproceedings{yang2022temporally,
  title={Temporally efficient vision transformer for video instance segmentation},
  author={Yang, Shusheng and Wang, Xinggang and Li, Yu and Fang, Yuxin and Fang, Jiemin and Liu, Wenyu and Zhao, Xun and Shan, Ying},
  booktitle={Proceedings of the IEEE/CVF Conference on Computer Vision and Pattern Recognition},
  pages={2885--2895},
  year={2022}
}

@inproceedings{jin2017video,
  title={Video scene parsing with predictive feature learning},
  author={Jin, Xiaojie and Li, Xin and Xiao, Huaxin and Shen, Xiaohui and Lin, Zhe and Yang, Jimei and Chen, Yunpeng and Dong, Jian and Liu, Luoqi and Jie, Zequn and others},
  booktitle={Proceedings of the IEEE International Conference on Computer Vision},
  pages={5580--5588},
  year={2017}
}

@inproceedings{gadde2017semantic,
  title={Semantic video cnns through representation warping},
  author={Gadde, Raghudeep and Jampani, Varun and Gehler, Peter V},
  booktitle={Proceedings of the IEEE International Conference on Computer Vision},
  pages={4453--4462},
  year={2017}
}

@inproceedings{nilsson2018semantic,
  title={Semantic video segmentation by gated recurrent flow propagation},
  author={Nilsson, David and Sminchisescu, Cristian},
  booktitle={Proceedings of the IEEE conference on computer vision and pattern recognition},
  pages={6819--6828},
  year={2018}
}

@article{yang2022decoupling,
  title={Decoupling features in hierarchical propagation for video object segmentation},
  author={Yang, Zongxin and Yang, Yi},
  journal={Advances in Neural Information Processing Systems},
  volume={35},
  pages={36324--36336},
  year={2022}
}

@inproceedings{zhang2020unsupervised,
  title={Unsupervised Feature Propagation for Fast Video Object Detection Using Generative Adversarial Networks},
  author={Zhang, Xuan and Han, Guangxing and He, Wenduo},
  booktitle={MultiMedia Modeling: 26th International Conference, MMM 2020, Daejeon, South Korea, January 5--8, 2020, Proceedings, Part I 26},
  pages={617--627},
  year={2020},
  organization={Springer}
}

@inproceedings{choudhuri2023context,
  title={Context-Aware Relative Object Queries To Unify Video Instance and Panoptic Segmentation},
  author={Choudhuri, Anwesa and Chowdhary, Girish and Schwing, Alexander G},
  booktitle={Proceedings of the IEEE/CVF Conference on Computer Vision and Pattern Recognition},
  pages={6377--6386},
  year={2023}
}

@article{he2022inspro,
  title={Inspro: Propagating instance query and proposal for online video instance segmentation},
  author={He, Fei and Zhang, Haoyang and Gao, Naiyu and Jia, Jian and Shan, Yanhu and Zhao, Xin and Huang, Kaiqi},
  journal={Advances in Neural Information Processing Systems},
  volume={35},
  pages={19370--19383},
  year={2022}
}

@inproceedings{wu2023onlinerefer,
  title={Onlinerefer: A simple online baseline for referring video object segmentation},
  author={Wu, Dongming and Wang, Tiancai and Zhang, Yuang and Zhang, Xiangyu and Shen, Jianbing},
  booktitle={Proceedings of the IEEE/CVF International Conference on Computer Vision},
  pages={2761--2770},
  year={2023}
}
